\documentclass[letterpaper]{article} %
\usepackage[draft]{aaai2026}  %
\usepackage{times}  %
\usepackage{helvet}  %
\usepackage{courier}  %
\usepackage[hyphens]{url}  %
\usepackage{graphicx} %
\usepackage{natbib}  %
\usepackage{caption} %
\usepackage{algorithm}
\usepackage{algorithmic}

\usepackage{newfloat}
\usepackage{listings}
\DeclareCaptionStyle{ruled}{labelfont=normalfont,labelsep=colon,strut=off} %
\floatstyle{ruled}
\newfloat{listing}{tb}{lst}{}
\floatname{listing}{Listing}

\usepackage{lineno}
\usepackage{tabularx}
\usepackage{graphicx}
\usepackage{caption}
\usepackage{subcaption}
\usepackage{color}
\usepackage{xcolor}
\usepackage{multirow}
\usepackage{tcolorbox}
\usepackage{amsmath}
\usepackage{textcomp}
\usepackage{soul}
\sethlcolor{white}
\usepackage{booktabs}
\usepackage{xurl}

\title{Framing Migration: A Computational Analysis of UK Parliamentary Discourse}
\author {
    Vahid Ghafouri\textsuperscript{\rm 1},
    Robert McNeil\textsuperscript{\rm 2},
    Teodor Yankov\textsuperscript{\rm 1},\\
    Madeleine Sumption\textsuperscript{\rm 2},
    Luc Rocher\textsuperscript{\rm 1},
    Scott A. Hale\textsuperscript{\rm 1,3},
    Adam Mahdi\textsuperscript{\rm 1},
}
\affiliations {
    \textsuperscript{\rm 1}Oxford Internet Institute, University of Oxford\\
    \textsuperscript{\rm 2}Centre on Migration, Policy and Society, University of Oxford\\
    \textsuperscript{\rm 3}Meedan\\
    vahid.ghafouri@oii.ox.ac.uk%
}

\begin{document}

\maketitle

\begin{abstract}
We present a large-scale computational analysis of migration-related discourse in UK parliamentary debates spanning over 75 years and compare it with US congressional discourse. Using open-weight LLMs, we annotate each statement with high-level stances toward migrants and track the net tone toward migrants across time and political parties. For the UK, we extend this with a semi-automated framework for extracting fine-grained narrative frames to capture nuances of migration discourse. Our findings show that, while US discourse has grown increasingly polarised, UK parliamentary attitudes remain relatively aligned across parties, with a persistent ideological gap between Labour and the Conservatives, reaching its most negative level in 2025. The analysis of \textit{narrative frames} in the UK parliamentary statements reveals a shift toward securitised narratives such as \textit{border control} and \textit{illegal immigration}, while longer-term integration-oriented frames such as \textit{social integration} have declined. Moreover, discussions of \textit{national law about immigration} have been replaced over time by \textit{international law} and \textit{human rights}, revealing nuances in discourse trends. Taken together broadly, our findings demonstrate how LLMs can support scalable, fine-grained discourse analysis in political and historical contexts.
\end{abstract}

\section{Introduction}
\label{sec:intro}

Immigration has long been one of the most divisive and fiercely contested topics in British society, shaping political discourse and public sentiment across decades. From the post-war arrivals of Commonwealth citizens to more recent discussions surrounding Brexit and the treatment of asylum seekers, migration has been a consistent feature in UK parliamentary and societal debates. Recent policies and policy proposals have included a May 2025 parliamentary white paper titled ``\textit{Restoring control over the immigration system}''\footnote{\url{www.gov.uk/government/publications/restoring-control-over-the-immigration-system-white-paper}} that proposed to double the path to British permanent residence to 10 years, the previous Conservative government's controversial Rwanda Scheme, the Cameron era `net migration target' and the Blair government's Immigration and Asylum Act 1999. 

Measuring how attitudes toward migrants have evolved over time is important for understanding policy shifts, anticipating policy and public opinion changes in reactions to events, and understanding agenda-setting effects~\cite{camargo2021agenda}.  Here, we analyse narratives expressed in
parliamentary debates on immigration, as Members of Parliament (MPs) not only reflect but also influence public sentiment~\cite{Bandeira2023representation, allen2019representation, ruhs2019representation}. 
Parliamentary data provides a multi-decade dataset of rich data on the national framings of immigration with metadata on political party, gender, age, and other attributes.

There remains a pressing need for automated quantitative approaches that capture both a high-level overview and fine-grained insights into migration-related discourse. While a high-level perspective enables the identification of broad patterns such as changes in framing over time or variation across political parties, lower-level and fine-grained analysis is essential for uncovering the specific narrative frames and rhetorical strategies that shape these patterns, including how different parties engage with issues like labour migration, asylum, and integration. Researchers have shown that classical NLP methods such as sentiment analysis and stance detection can help explore political discourse. For example, \citet{Card2022USDebates} and \citet{kostikova2024germanparliament} examined migration-related speech in the US and German parliaments, respectively. However, these studies are limited in both their granularity and scope. UK parliamentary discourse remains underexplored, and previous methods tend to miss subtle rhetorical variation. Until now, performing a systematic cross-parliamentary comparison of migration discourse across multiple national legislatures has remained challenging.

Recent advancements in large language models (LLMs), however, present an opportunity to move beyond these limitations by enabling more granular analyses of complex social phenomena. Emerging work in this area has demonstrated the potential of LLMs to generate deeper contextual understanding in both high-level and fine-grained social computing tasks~\cite{cetinkaya2025cpi, zhu2024chatgpt}.

This paper contributes to both methods and applications. We first build a novel dataset of UK parliamentary statements with comprehensive author-level metadata,  through web crawling and retrieval-augmented generation (RAG). We then extract rich semantic metadata from statements, including stances toward migrants and narrative frames, using lightweight LLMs. We analyse their associations with parameters such as \textit{time}, \textit{party}, \textit{gender}, and \textit{age}. We also introduce a systematic pipeline for discovering and extracting narrative frames through iterative interaction between LLMs and human feedback. Overall, we present the first large-scale, automated analysis of migration discourse in UK parliamentary debates and conduct a high-level comparative analysis with US congressional debates.

Our analysis reveals notable patterns: UK parliamentary attitudes toward migrants, while ideologically split, remain more convergent across parties than in the US, with Labour consistently more solidarity-oriented than the Conservatives. We also observe a recent shift toward securitised rhetoric, such as ``\textit{illegal immigration}'' and ``\textit{border control}'', often at the expense of longer-term integration themes. Additionally, many \textit{anti-solidarity} statements toward migrants appear in mixed or ambiguous form, reflecting indirect rhetorical strategies shaped by enduring antipopulist norms of political decorum in the UK, compared to the US~\cite{freeman1995antipopulist}.

\section{Method}
\label{sec:method}

\subsection{Data Curation}
\label{sec:data}

\subsubsection{Data Collection}
\label{sec:data:collection}

We collected UK parliamentary debate data (House of Commons and Lords) from 1800 until July 2025, using the Hansard API\footnote{\url{https://hansard-api.parliament.uk/}}, a public API that provides access to transcribed statements from UK parliamentary proceedings. We collected US parliamentary data (Congress and Senate), by combining data from \citet{Card2022USDebates}, spanning 1870 to 2020, with newer data from 2020 to July 2025 using the US Congressional Record API\footnote{\url{https://github.com/unitedstates/congressional-record/}}. To keep the focus on migration-related discourse, we queried the API for statements containing keywords ``\textit{refugee(s)}'', ``\textit{migrant(s)}'', ``\textit{immigrant(s)}'', ``\textit{asylum}'', ``\textit{migration}'', and ``\textit{immigration}''.

In total, we collected $118,603$ statements from the UK Commons and Lords records, as well as $153,523$ statements from the US Congress and Senate parliamentary records.
Each statement contains multiple attributes in both APIs, from which we use the following: the \textit{full text}, for semantic analyses, and the \textit{sitting date} (the time of the parliamentary meeting), for adding a temporal element to the analyses. The main MP attributes we use in this paper are \textit{Party}, \textit{Gender}, and \textit{Birth Date} (subtraction of each statement's sitting date and MP's Birth Date provides us with the MP's age at the time of making the statement).

Unlike US data that already includes speaker-level info of Congresspeople/Senators (e.g. ``party''), the Hansard API only provided the ``\textit{speaker IDs}'' and ``\textit{honorary titles}'' of the speakers (e.g., ``\textit{Lord Hunt of Kings Heath}'') in the UK parliament. Moreover, unlike the ``\textit{honorary titles}'' that were well-recorded in the dataset, most of the ``\textit{speaker IDs}'' were missing for pre-2005 statements. Thus, to collect the mentioned speaker-level info for the UK dataset, we searched the British MPs' \textit{honorary titles} online, and extracted the desired metadata from their Wikipedia page using Retrieval Augmented Generation (RAG). See Tables~\ref{tab:contributions} and~\ref{tab:mp_metadata} for the full list of columns in the dataset.

\begin{table*}[h]
    \centering
    \begin{tabular}{p{3cm}|p{13.5cm}}
        \toprule
        \textbf{Field} & \textbf{Description} \\
        \midrule
        Body             & Full text of the statement contribution. We will refer to it as a ``\textit{statement}'' throughout this paper.\\
        Member ID         & Unique identifier for the speaker. (Unfortunately, it was missing for pre-2005 data, which is why we used the \textit{MP's Honorary Title} to collect the corresponding member's data from the web. \\
        Sitting Date      & Date and time when the contribution was made. \\
        House            & Legislative body where the contribution occurred (e.g., House of Commons, House of Lords, Westminster Hall, Scottish Parliament). \\
        MP's Honorary Title & Official title by which the MP is addressed in parliament (e.g., \textit{Lord Hunt of Kings Heath}). \\
        \bottomrule
    \end{tabular}
    \caption{Key fields in the parliamentary statements' dataset.}
    \label{tab:contributions}
\end{table*}

\begin{table*}[h]
    \centering
    \begin{tabular}{p{3.5cm}|p{13cm}}
        \toprule
        \textbf{Field} & \textbf{Description} \\
        \midrule
        MP's Attribution      & Official name and title of the member (e.g., \textit{Lord Hunt of Kings Heath}). \\
        Gender                & Gender identity of the member. \\
        Birth Date           & Date of birth of the member. \\
        Birth City           & City where the member was born. \\
        Party                 & Political party affiliation at the time of the contribution. \\
        House                 & Legislative chamber where the member served. \\
        House Start Date    & Start date of the member’s service in the given house. \\
        House End Date      & End date of the member’s service in the given house. \\
        Constituency Country & Country of the member’s constituency (e.g., England, Scotland, Wales, Northern Ireland). \\
        Constituency Region  & Name of the member’s constituency or representative region. \\
        \bottomrule
    \end{tabular}
    \caption{Metadata fields for members of parliament.}
    \label{tab:mp_metadata}
\end{table*}

Since the members' complete metadata was largely missing from the Hansard database, we employed a Retrieval Augmented Generation (RAG) pipeline using the LangChain library to extract it. First, we located each MP’s exact Wikipedia page by querying their parliamentary attribution (e.g., \textit{Lord Browne of Ladyton}) using the DuckDuckGo search API in Python. We then retrieved the full content of the corresponding Wikipedia page via the Wikipedia API integrated in LangChain. Finally, we extracted structured metadata from the retrieved text using a lightweight, instruction-tuned language model (\textit{Mistral-7B-Instruct-v0.2}).

\noindent \textbf{Note:} As the pre-1950 records were sparse in both UK and US datasets (less than 10\% of the total for each), for \textit{temporal analyses} in Figure~\ref{fig:party_time_attitude_comparison}~\ref{fig:frames_temporal}, we only portray results for 1950 until 2025 July.

We will release the full dataset and code on GitHub in the Camera-Ready version.

\subsubsection{Data Annotation}
\label{sec:data:annotation}

We prompt LLMs to generate both high-level and fine-grained annotations for each statement. The \textit{high-level stances} and \textit{narrative frames} annotations, explained as follows, correspond to the two mentioned levels of granularity.

Our main LLM of choice throughout the paper is \textit{Qwen3-32B}\footnote{\url{https://huggingface.co/Qwen/Qwen3-32B}}, which, at the time of carrying out the experiments, ranked as the state-of-the-art\footnote{\url{https://lmarena.ai/leaderboard}} among open-source LLMs under 50B parameters. For validation and comparison, we also utilise \textit{Mistral-7B-Instruct-v0.2}\footnote{\url{https://huggingface.co/mistralai/Mistral-7B-Instruct-v0.2}}, a strong and efficient lightweight model previously adopted in computational social science research. The default values are used for all hyperparameters. All experiments are run on an \textit{NVIDIA L40S} GPU. With 4-bit quantisation, inference times per prompt average approximately 0.7 seconds for \textit{Mistral-7B-Instruct-v0.2} and 3 seconds for \textit{Qwen3-32B}.

\noindent \textbf{Relevance Filtering:}
As our data collection step relies on keyword search, our initial annotation step is to filter out statements that are irrelevant to migration to the UK (US) from the analysis. For instance, we have encountered UK parliamentary texts containing migration-related keywords, yet they merely discuss the migration of Palestinians to Egypt and Jordan. We consider that this is outside our scope. Thus, we initially tasked \textit{Qwen3-32B} to classify statements as \textit{relevant} or \textit{irrelevant} to migration to the UK (US). See Table~\ref{tab:prompt_relevance} in the Appendix~\ref{sec:appendix:prompts} for the prompt. This step, filtered out 14.7\% (3.4\%) of the statements from the UK (US) dataset.

\noindent \textbf{High-level Stances:}  
Following and adapting the categorisation of \citet{thijssen2012solidarity} in social science, which is also adopted by \citet{kostikova2024germanparliament} in NLP literature, we classify each statement into one of four high-level categories: \textit{solidarity}, \textit{anti-solidarity}, \textit{mixed}, or \textit{none}. We ask LLM to label a statement as \textit{solidarity} if it expresses support, empathy, or compassion toward immigrants. In contrast, we instruct LLM to label statements as \textit{anti-solidarity} if they advocate opposition, exclusion, or restriction toward migrants. We define the \textit{mixed} category as statements that contain both supportive and opposing expressions in the same text--e.g., recognising migrants’ rights while limiting their support. Finally, we assign the \textit{none} label to statements that are off-topic or contain no discernible stance toward migrants. This four-way categorisation allows us to distinguish between overt positions, ambivalent rhetoric, and neutral discourse. The full LLM prompt used to elicit these labels is provided in Appendix Table~\ref{tab:prompt_solidarity}.
While our prompt includes an option for specifying subtypes of solidarity and anti-solidarity, we do not use these subcategories in the present analysis, as we introduce our own approach to increasing the granularity.

\noindent \textbf{Disambiguating Mixed Stances:}  
While we design the \textit{mixed} label to capture rhetorical ambivalence---statements that balance supportive and restrictive tones---it often includes cases where the overall policy intent is eventually aligned with either \textit{solidarity} or \textit{anti-solidarity}. To better capture this distinction, we implement a second-pass annotation procedure specifically for statements initially labelled as \textit{mixed}. In this step, we prompt the LLM (Qwen3-32B) to assess the \textit{final intent} or \textit{policy direction} of the statement, going beyond surface-level tone to evaluate whether the speaker ultimately supports or seeks to restrict the rights or social standing of migrants.

The full prompt for this second-pass reclassification is provided in Appendix Table~\ref{tab:prompt_mixed}. The output consists of three labels: \textit{solidarity}, \textit{anti-solidarity}, or \textit{none}; where the model is not provided with the option of \textit{mixed} to be forced to disambiguate the statements.

For instance, the following statement was initially classified as \textit{mixed} by both \textit{Mistral-7B} and \textit{Qwen3-32B}, but re-labelled as \textit{anti-solidarity} after applying the disambiguation prompt:

\begin{quote}
\label{quote:mixed-sample}
\small
``\textit{Is the Undersecretary aware that most reasonable people want to see a fair deal given to those immigrants who are already here and to consolidate their position, and that what the people are concerned about is the tremendous number of dependants who will come in in the years ahead? They must be limited if those already here are to get a square deal.}''\footnote{\url{https://hansard.parliament.uk/Commons/1968-07-02/debates/18740306-4db5-43c3-84c2-d62c58fdb678/Immigrants}}
\end{quote}

While the statement frames its initial argument in terms of fairness toward current migrants, it ultimately justifies restricting the future migration of their dependants, revealing a restrictive policy stance. This habit might ambiguate the results when computing the overall tone, possibly underestimating the rate of \textit{anti-solidarity} discourse, especially in parliamentary environments where anti-populist norms~\cite{freeman1995antipopulist} encourage the use of \textit{mixed} tones for \textit{anti-solidarity} intents.
This revision process allows us to reduce ambiguity in the \textit{mixed} category and more accurately reflect the underlying intention of each contribution. Throughout Section~\ref{sec:results}, we mainly use the post-disambiguation annotations for the analyses, except for Figure~\ref{fig:labels-with-mixed}, where we report the rate of \textit{mixed} labels for its unique insights.

\noindent \textbf{Narrative Frames:}
To increase the granularity of the textual analysis, we extract \textit{narrative frames} that characterise the rhetorical context in which migration is discussed. We employ a semi-automated narrative frame extraction pipeline that combines human domain expertise with LLM-assisted labelling in an iterative feedback loop.

\begin{figure}[ht]
    \centering
    \includegraphics[width=\linewidth]{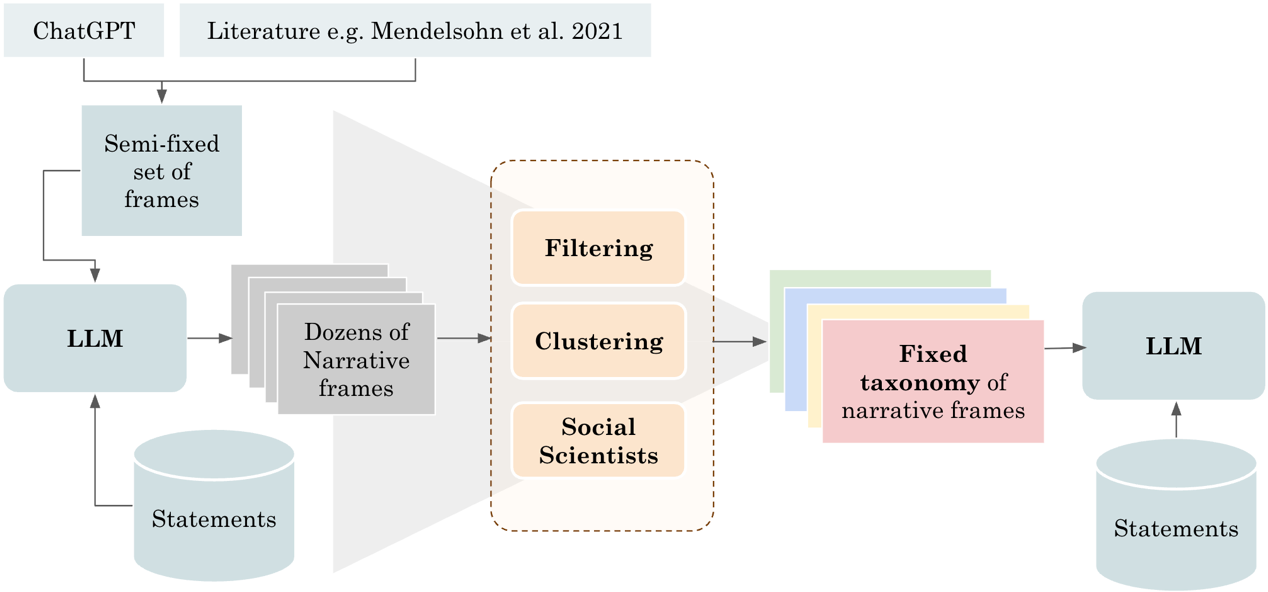}
    \caption{Narrative Frame Extraction Pipeline. We start with a seed taxonomy of narrative frames about migrants, use LLM for semi-bounded frame classification with label expansion, and finalise the taxonomy through iterative expert review and consolidation. Finally, we ask the LLM to annotate all the statements from the fixed seed of narrative frames. Figure~\ref{fig:wordcloud_evolution} in Appendix~\ref{sec:appendix:narrative-frames} portrays the list and evolution of the taxonomy.}
    \label{fig:narrative_pipeline}
\end{figure}

Figure~\ref{fig:narrative_pipeline} shows the pipeline. We first curated an initial taxonomy using the taxonomy of migration-related narrative frames (e.g., \textit{humanitarian obligation toward migrants}, \textit{ migrants' economic contribution}, \textit{border control}) from \citet{mendelsohn2021frame} and additional suggestions from \textit{GPT-4o}. We then systemically and iteratively expanded the taxonomy into a fixed list of 61 comprehensive narrative frames. The full initial and final list of the taxonomy is visible in Figure~\ref{fig:wordcloud_evolution} in Appendix~\ref{sec:appendix:narrative-frames}. Also, the narrative extraction prompt in Table~\ref{tab:prompt_frames_bounded} lists the full narrative frames taxonomy.

\subsection{Statistical Analyses}
\label{sec:method:association}

To identify temporal and political trends in migration discourse, we conduct statistical analyses between different dimensions of our annotations. The choice of analytical method depends on the nature of the variables: correlation analysis for continuous variables, cross-tabular analysis for categorical variables, and chi-square tests for independence testing.

\subsubsection{Net Tone Over Time}
\label{sec:method:association:net-tone}

To examine how overall attitudes toward immigrants vary across parties and over time, we compute a net tone measure that captures the balance between \textit{solidarity} and \textit{anti-solidarity} discourse.

Following \citet{Card2022USDebates}, we calculate the net tone using the \textit{High-level Stance} annotations from Section~\ref{sec:data:annotation}:

\begin{equation} \label{eq:net-tone} \textit{net tone} = \frac{N_{\text{solidarity}} - N_{\text{anti-solidarity}}}{N_{\text{solidarity}}+N_{\text{anti-...}}+N_{\text{none}}} \end{equation}

where $N_{\text{solidarity}}$ and $N_{\text{anti-solidarity}}$ are the counts of solidarity and anti-solidarity statements, respectively, and $N_{\text{total}}$ is the total number of statements. This measure ranges from -1 (utter \textit{anti-solidarity}) to +1 (utter \textit{solidarity}), with 0 indicating a balanced discourse.

We compute this measure across different temporal aggregations (e.g., yearly and by government period) and stratify by political party, gender, and age to identify systematic patterns in migration discourse.

\subsubsection{Narrative Frame Associations} \label{sec:method:association:narrative}

To understand how specific narrative frames are associated with different stances toward migration, we analyse the relationship between our predefined narrative frames and high-level solidarity classifications.

Since both variables (frame and stance) are categorical, we employ chi-square tests of independence to identify frames that are significantly associated with particular stances. For each frame $f$, we construct a $2 \times 2$ contingency table comparing its frequency of use in \textit{solidarity} versus \textit{anti-solidarity} contexts:

\begin{equation} \chi^2_f = \sum_{i,j} \frac{(O_{ij} - E_{ij})^2}{E_{ij}} \end{equation}

where $O_{ij}$ represents the observed frequency of frame $f$ in stance category $j$ (solidarity or anti-solidarity), and $E_{ij}$ is the expected frequency under the null hypothesis of independence.

To ensure statistical robustness, we apply a frequency filter, including only frames that appear in at least 2,500 total statements. We use a significance threshold of $p < 0.01$ to identify frames with strong associations to particular stances. For each significant frame, we calculate the difference in proportional use between \textit{solidarity} and \textit{anti-solidarity} contexts to quantify the direction and magnitude of association.

\section{Results}
\label{sec:results}

\subsection{High-level Attitudes}
\label{sec:results:high-level-attitudes}

\begin{figure}[htbp]
    \centering
    \begin{subfigure}[t]{0.48\columnwidth}
        \centering
        \includegraphics[width=\textwidth]{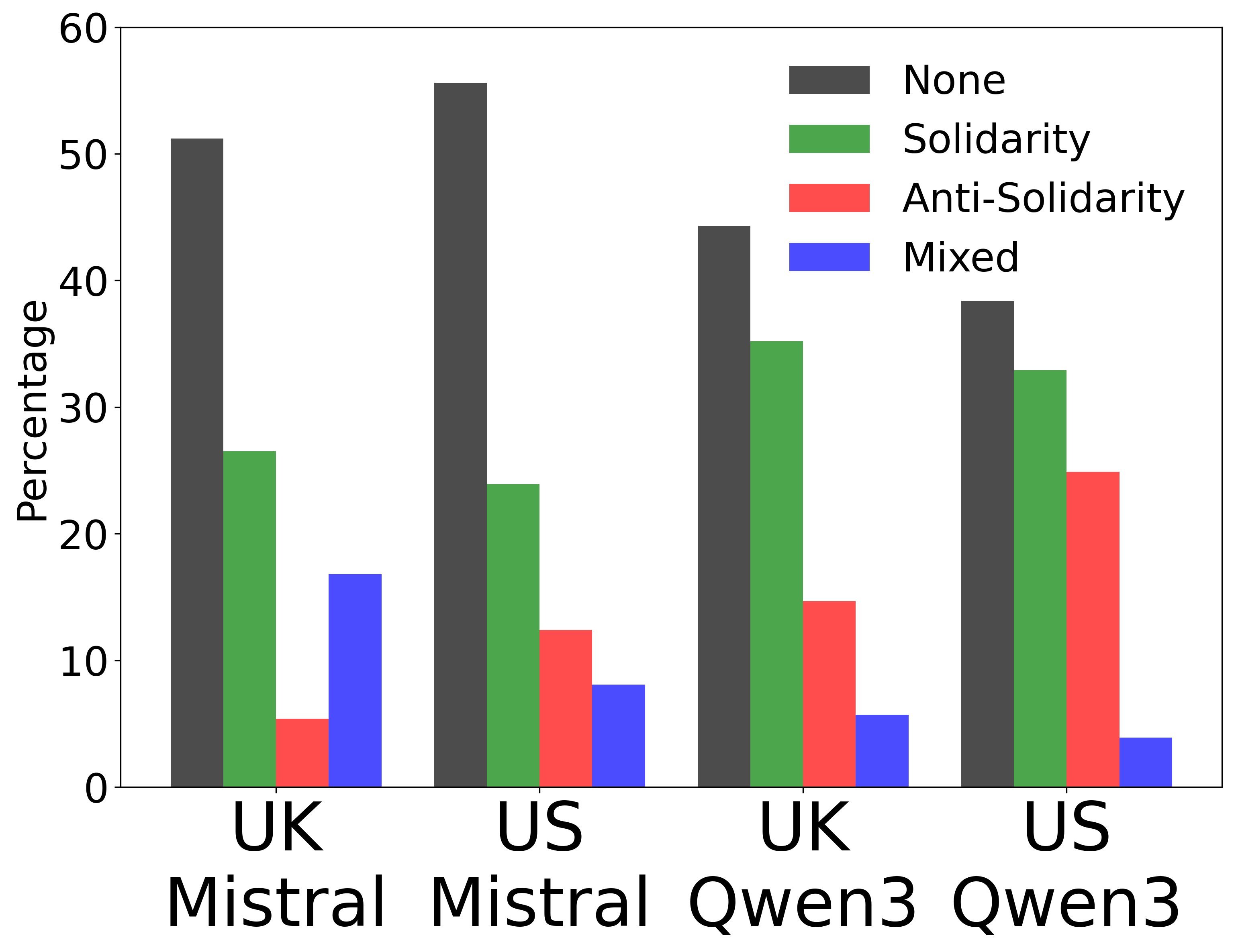}
        \caption{With \textit{mixed} class}
        \label{fig:labels-with-mixed}
    \end{subfigure}%
    \hfill
    \begin{subfigure}[t]{0.48\columnwidth}
        \centering
        \includegraphics[width=\textwidth]{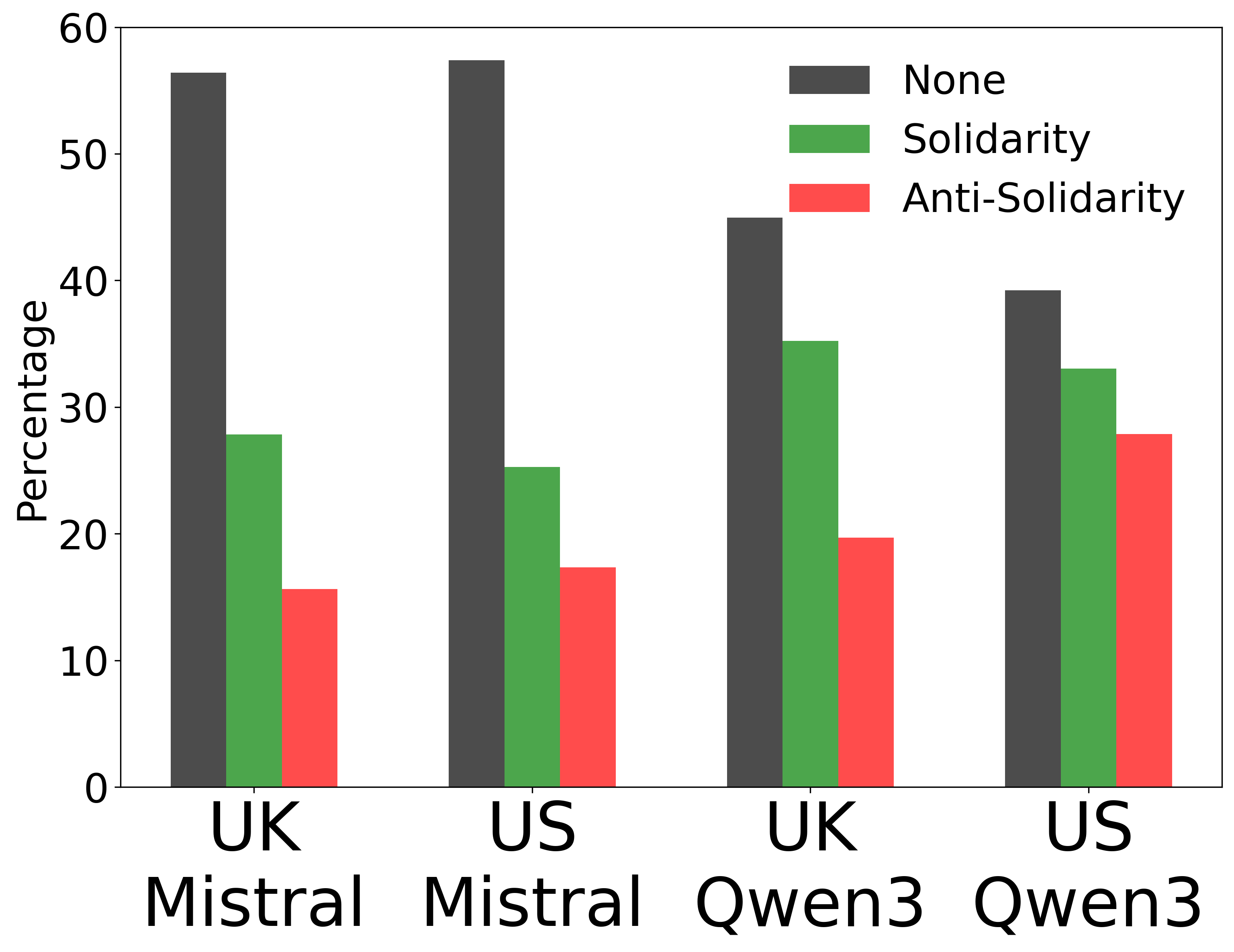}
        \caption{\textit{mixed} class redistributed}
        \label{fig:labels-aggregated}
    \end{subfigure}
    \caption{Solidarity Classification Results Across Datasets. The chart shows the distribution of solidarity categories (\textit{none}, \textit{solidarity}, \textit{anti-solidarity}, and \textit{mixed}) as classified by two LLMs: \textit{Mistral-7B-Instruct-v0.2} and \textit{Qwen3-32B}. Regarding \textit{mixed} statements (statements with ambiguous mixtures of \textit{solidarity} and \textit{anti-solidarity} stances, Figure~\ref{fig:labels-with-mixed} shows them as a separate class while Figure~\ref{fig:labels-aggregated} redistributes them to other classes after the disambiguation prompt (Table~\ref{tab:prompt_mixed}), which uncovers the underlying intent of the \textit{mixed} statements; most of which end up classified as \textit{anti-solidarity}.}
    \label{fig:labels-breakdown}
\end{figure}

Figure~\ref{fig:labels-breakdown} portrays the breakdown of high-level attitude labels for the UK and US datasets for both pre-disambiguation (Figure~\ref{fig:labels-with-mixed}) and post-disambiguation (Figure~\ref{fig:labels-aggregated} phase.

\begin{figure*}[h]
    \centering
    \begin{subfigure}[t]{0.48\linewidth}
        \centering
        \includegraphics[width=\linewidth]{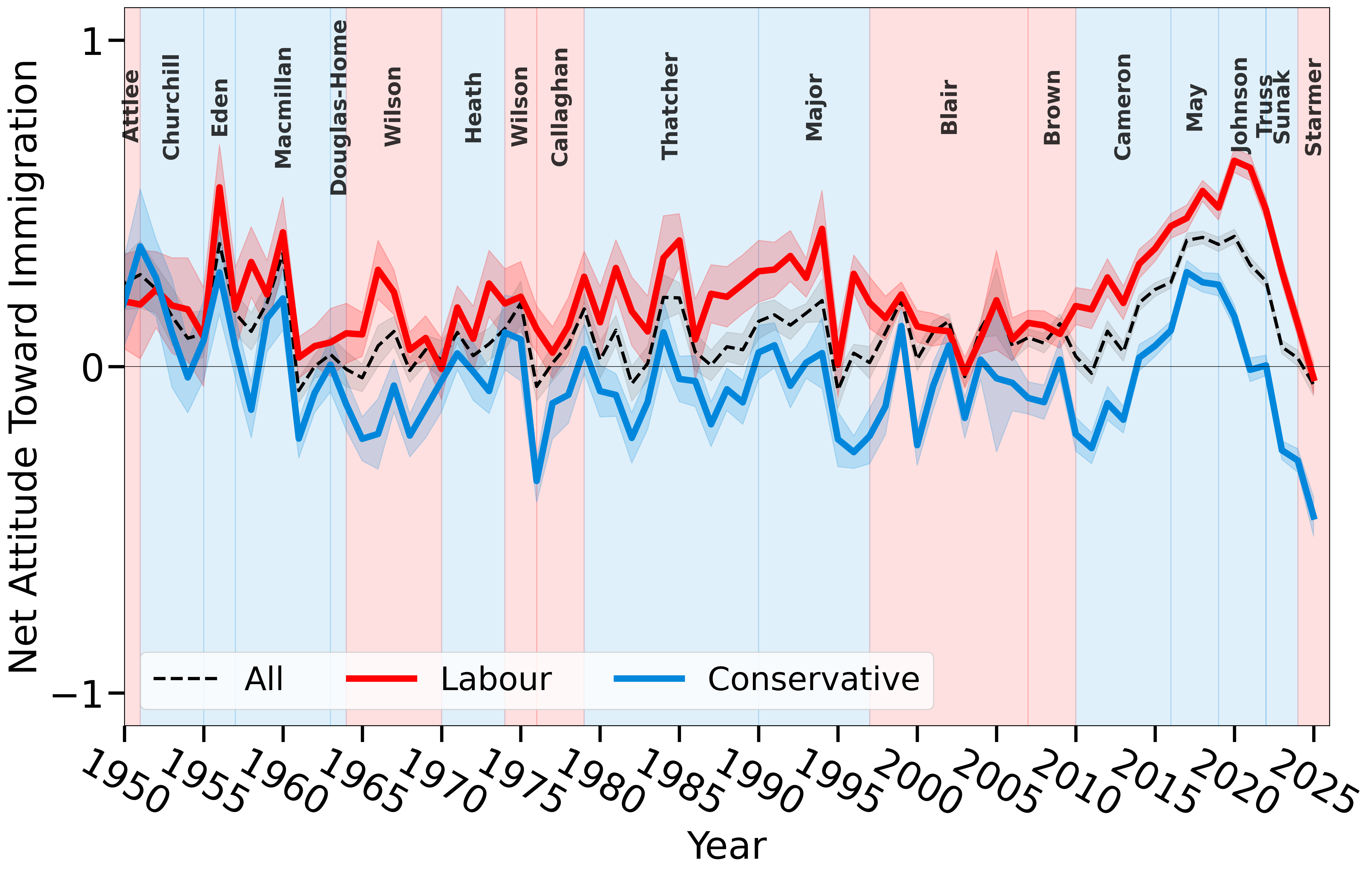}
        \caption{UK}
        \label{fig:uk_party_attitude}
    \end{subfigure}
    \hfill
    \begin{subfigure}[t]{0.48\linewidth}
        \centering
        \includegraphics[width=\linewidth]{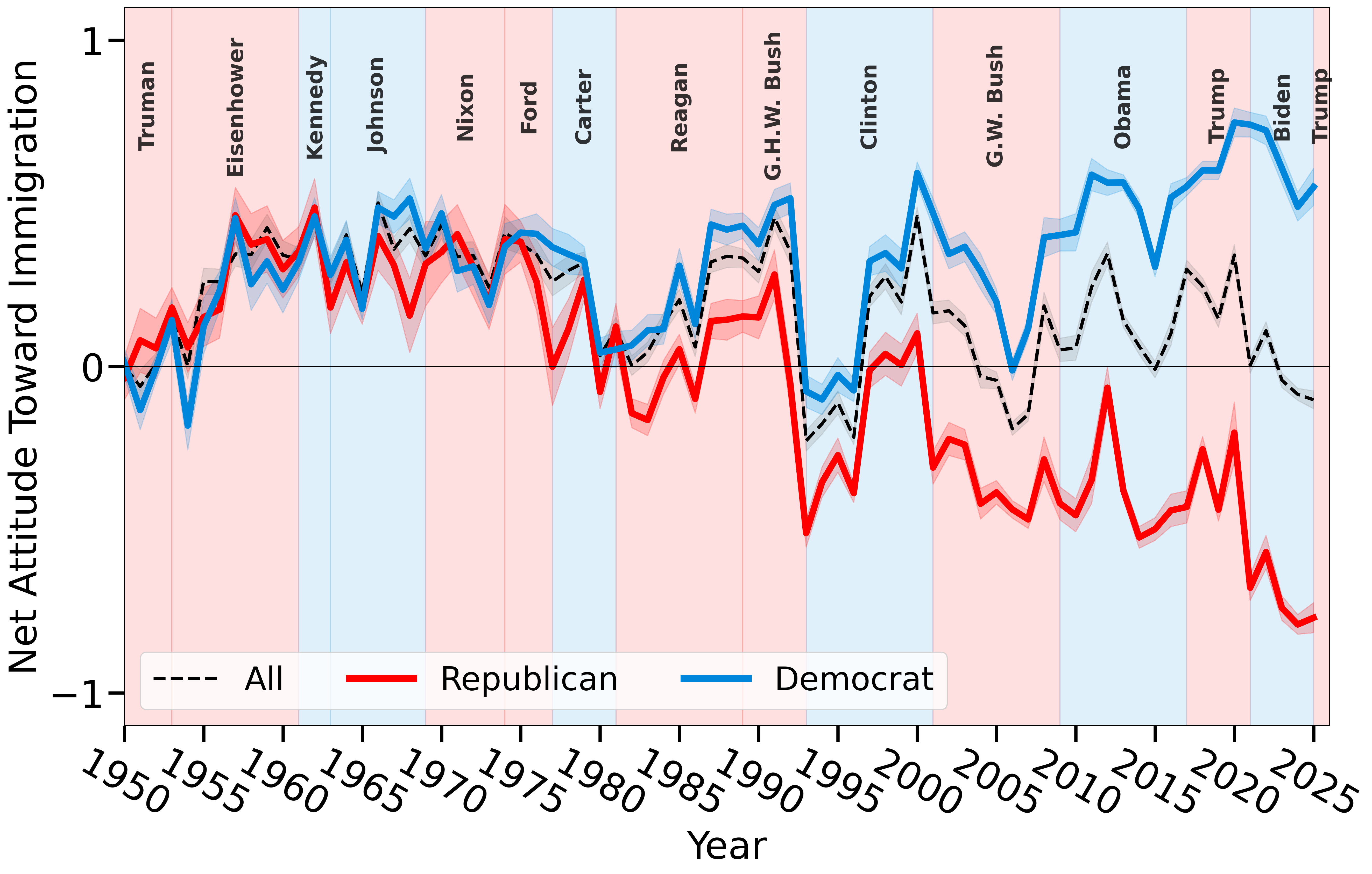}
        \caption{US}
        \label{fig:us_party_attitude}
    \end{subfigure}
    \caption{\textit{Net attitude towards immigration.} The y-axis shows net attitude from Equation~\ref{eq:net-tone} per year. The dashed line shows the average, with bands indicating $\pm$2 standard deviations. The Spearman correlation of the net attitudes between parties across years is $45\%$ in the UK and $-11\%$ in the US, suggesting a higher party alignment in the UK and higher polarisation in the US.}
\label{fig:party_time_attitude_comparison}
\end{figure*}

Further, the temporal results for the UK and US are visualised in Figure~\ref{fig:party_time_attitude_comparison}, which plots the average tone toward immigrants over time for left-leaning (Labour, Democrat) and right-leaning (Conservative, Republican) parties based on Equation~\ref{eq:net-tone}.

\begin{figure}[ht]
\centering \includegraphics[width=\linewidth]{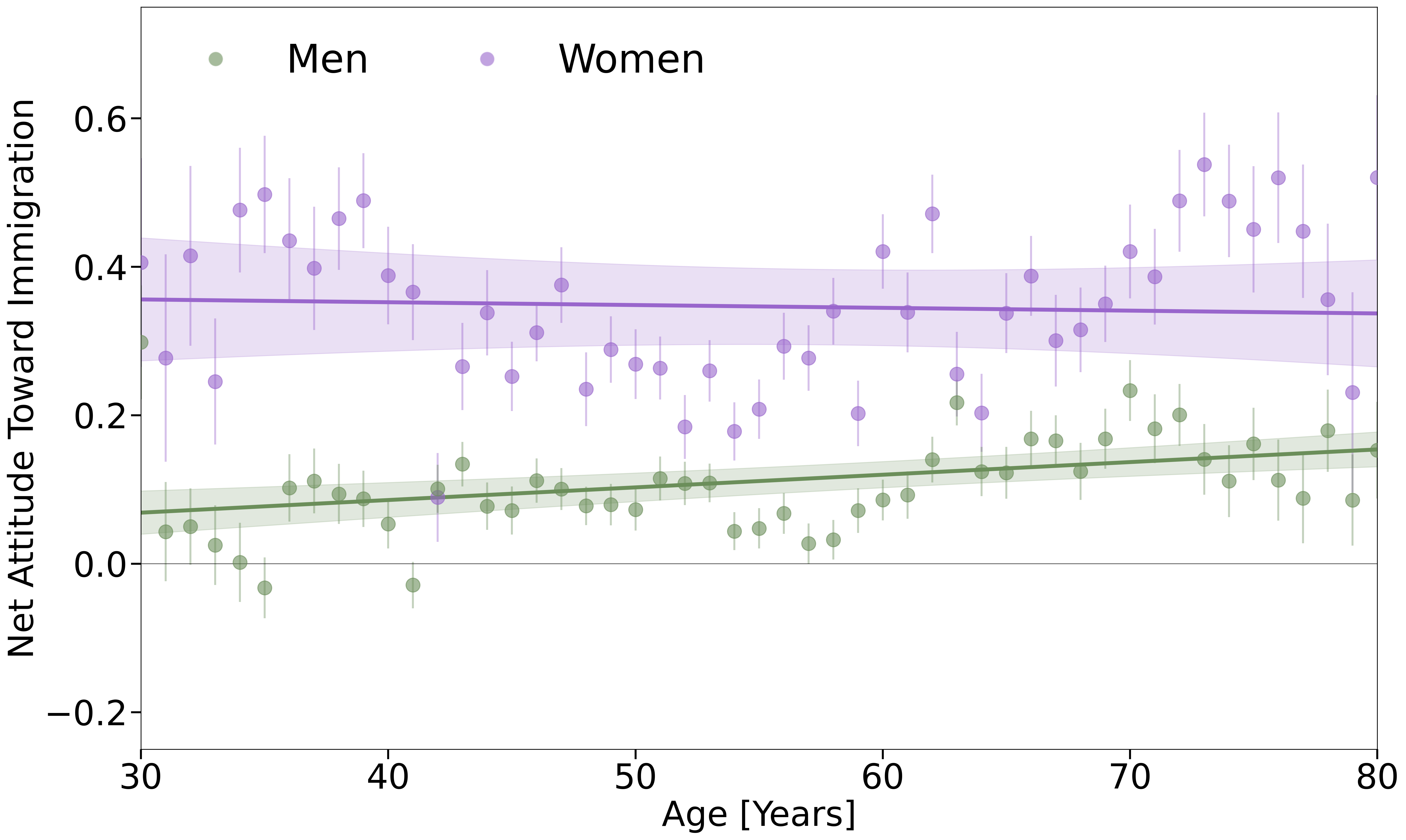}
\caption{Association between net attitude toward immigrants (y-axis based on Equation~\ref{eq:net-tone}) with gender and age of UK MPs. Error bars indicate $\pm$2 standard deviations. Solid lines show linear regression fits with 95\% confidence intervals (shaded regions).}
\label{fig:age_gender_attitude}
\end{figure}

Finally, Figure~\ref{fig:age_gender_attitude} explores the association between the UK MPs' age and gender vs. their net attitude toward migrants.

\subsection{Narrative frames}
\label{sec:results:frames}

\subsubsection{Frames' context}
\label{sec:results:frames:context}

\begin{figure}[t]
\centering
\includegraphics[width=\linewidth]{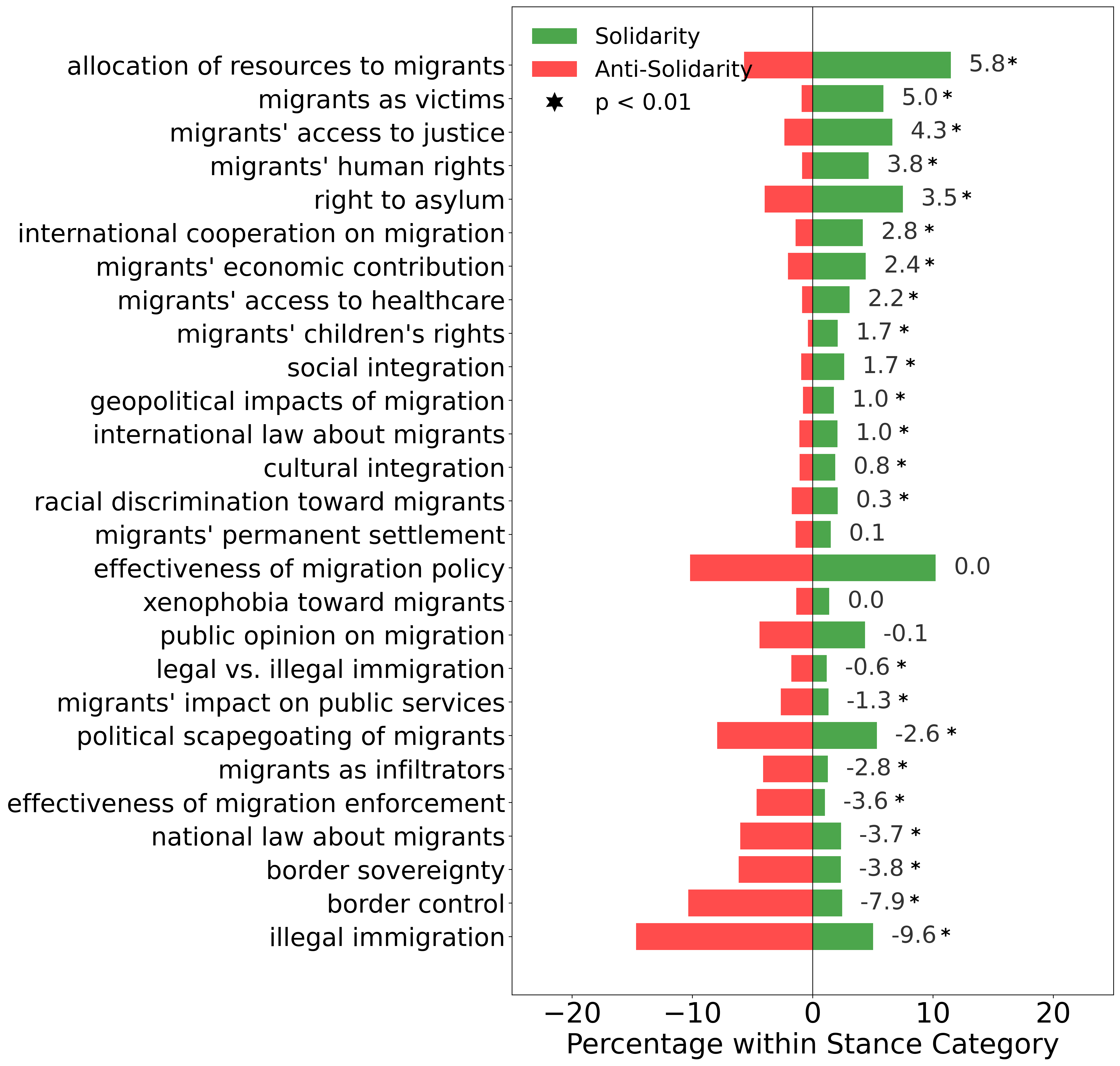} \caption{Diverging bar chart of narrative frame usage by political stance. Bars extend left (red) for \textit{anti-solidarity} and right (green) for \textit{solidarity} discourse, showing within-stance frequency percentages for frames. Frames are ranked by difference in usage rates between stances, from most \textit{anti-solidarity}-associated to most \textit{solidarity}-associated. Numbers indicate percentage point differences; stars denote statistically significant associations ($\chi^2$ test, $p < 0.01$).}
\label{fig:frames-context}
\end{figure}

Learning the context (\textit{solidarity} vs. \textit{anti-solidarity}) in which frames have been used could hint at the argumentative strategies of each side of the migration debate. Thus, we analysed the distribution of narrative frames within \textit{solidarity} and \textit{anti-solidarity} discourse using the methodology described in Section~\ref{sec:method:association:narrative}. Figure~\ref{fig:frames-context} presents a diverging bar chart showing the relative usage of narrative frames that meet our frequency threshold ($\geq2500$ total uses), calculated as within-stance percentages to account for class imbalance in our dataset.

From Figure~\ref{fig:frames-context}, we can observe that the main salient \textit{anti-solidarity} frames are the ones discussing \textit{illegal immigration} and potential contributing factors of it, such as \textit{border control} and \textit{border sovereignty}. Conversely, narrative frames that associate with migrants' rights and welfare, such as \textit{allocation of resources to migrants} and their \textit{human rights}, are the most salient in \textit{solidarity} discourse. Conversations about \textit{cultural}, \textit{social}, and \textit{economic} integration of migrants appear as less salient in the parliamentary debates.

\subsubsection{Frames' temporal evolution}
\label{sec:results:frames:temp}

\begin{figure*}[h]
\centering
\begin{subfigure}[b]{0.48\textwidth}
    \centering
    \includegraphics[width=\textwidth]{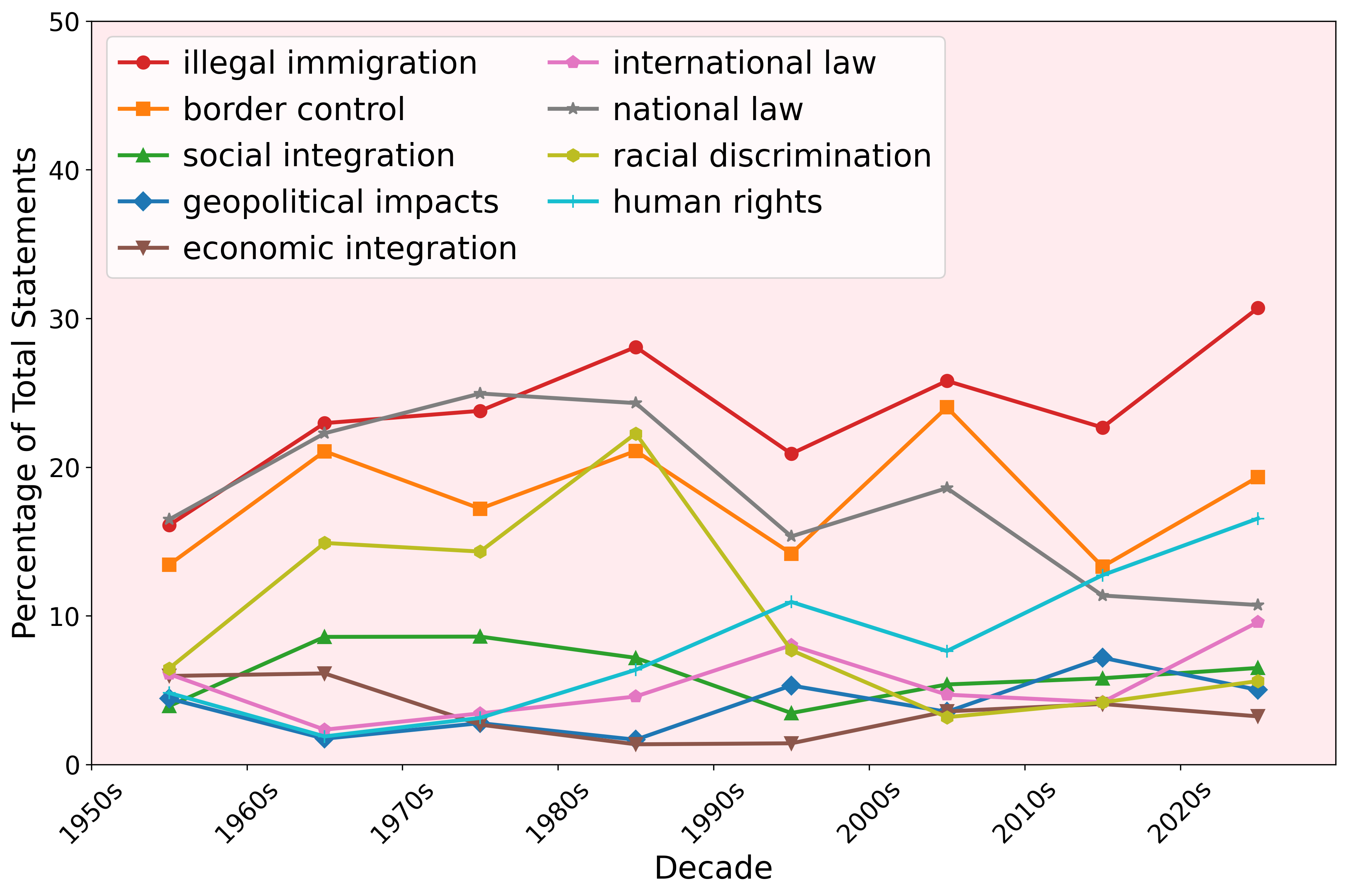}
    \caption{Labour Party}
    \label{fig:frames_temporal_labour}
\end{subfigure}
\hfill
\begin{subfigure}[b]{0.48\textwidth}
    \centering
    \includegraphics[width=\textwidth]{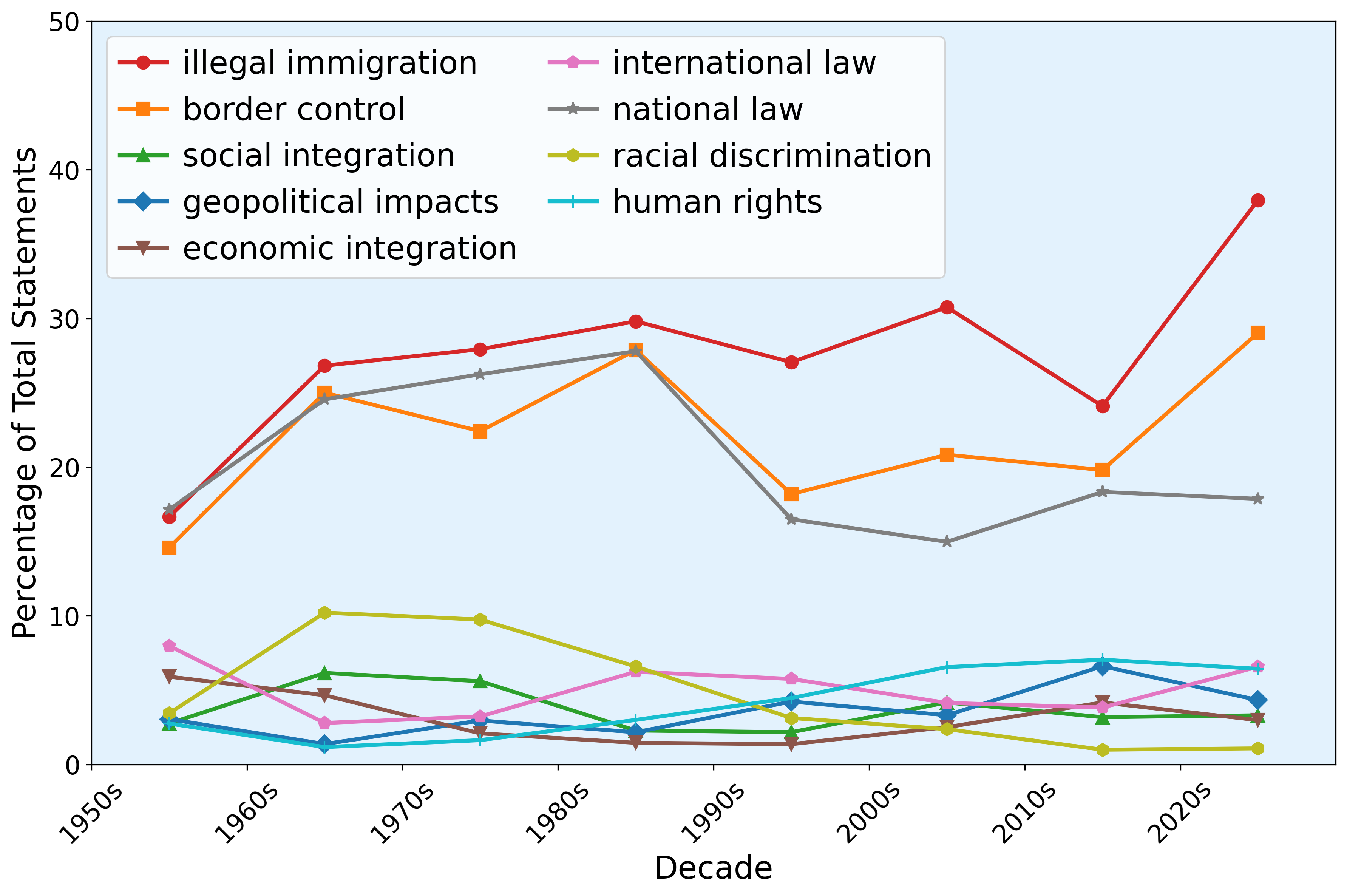}
    \caption{Conservative Party}
    \label{fig:frames_temporal_conservative}
\end{subfigure}
\caption{Temporal evolution of key narrative frames by political party, 1950s--2020s (until July 2025). Each line represents the percentage of total parliamentary contributions within each decade that employed the specified frame. Legend labels are shortened for space (e.g., ``\textit{geopolitical impacts}'' for ``\textit{geopolitical impacts of migration}'').}
\label{fig:frames_temporal}
\end{figure*}

We visualise the temporal distribution of a subset of key frames across decades from 1950 onwards. Figure~\ref{fig:frames_temporal} presents the percentage usage of nine prominent narrative frames for Labour and Conservative parties separately, calculated as the proportion of total parliamentary contributions within each decade that employed each frame.

As portrayed in Figure~\ref{fig:frames_temporal}, ``\textit{illegal immigration}'' and ``\textit{border control}'' are the most prominent frames for the whole time series, receiving an extra rise in the 2020s, especially for the Conservatives.

While both parties employ ``\textit{border control}'' framing, the Conservative Party shows consistently higher use across most decades, with this frame becoming increasingly prominent from the 1980s onwards. The Labour Party's use of this frame remains relatively stable but lower throughout the period. It witnessed a peak during the 2000s---a period when the Blair/Brown administration was dealing with record levels of asylum claims, unauthorised arrivals from Calais, the closure of the Sangatte refugee camp and the introduction of juxtaposed border controls.

The discussions of ``\textit{migrants' human rights}'' show partisan differences. Labour contributions consistently employ this frame at increasingly higher rates in the 2010s and 2020s, reaching peaks of approximately 15\% in recent decades. Conservative use of human rights framing remains notably lower and more stable across time. Labour's increasing focus on human rights frames is almost a mirror image of its reducing focus on national law. 

Frames related to integration of migrants, whether ``\textit{social integration}'' or ``\textit{economic integration}'', show relatively modest and stable (5\%--10\%) usage by both parties throughout the period. Yet, Labour shows slightly higher engagement with ``\textit{social integration}'' discourse, while both parties show similar patterns for ``\textit{economic integration}''. Notably, there had been a high emphasis on ``\textit{racial discrimination toward migrants}'' pre-1990s, especially by the Labour party in the 1980s. However, it turned into a minor issue for both parties since the 1990s.

\section{Validation}
\label{sec:validation}

To evaluate the reliability of our automated solidarity classification, we measure inter-annotator agreement between two LLMs (\textit{Mistral-7B-Instruct-v0.2} and \textit{Qwen3-32B}) and two human annotators who were co-authors. Each annotator was provided with 100 random samples and the same instructions as LLM prompts (see Appendix~\ref{sec:appendix:prompts}). Their demographics are not disclosed for anonymity.

Given the inherent subjectivity of labelling political discourse~\cite{ghafouri2023llmbias}, we report full-label agreement across four categories--\textit{solidarity}, \textit{anti-solidarity}, \textit{mixed} (before the disambiguation step), \textit{none}--and also binary agreement between \textit{solidarity} and \textit{anti-solidarity}.

The reason for reporting the binary agreement is that while we expect some variation in where the annotators draw the boundary between \textit{none}/\textit{mixed} vs. partisan labels (i.e. \textit{solidarity} and \textit{anti-solidarity}), our primary concern is avoiding critical misclassification between the partisan labels---where a statement expressing \textit{solidarity} is incorrectly labelled as \textit{anti-solidarity}, or vice versa. These \textit{fatal errors} could critically distort downstream analyses, leading to incorrect insights.

\begin{table}[h]
\centering
\small
\begin{tabular}{lcc}
\toprule
\textbf{Annotator Pair} & \textbf{4-Way} & \textbf{Binary} \\
\midrule
Mistral-7B vs Qwen3-32B & $\kappa = 0.49$ & $\kappa = 0.93$ \\
Mistral-7B vs Human A & $\kappa = 0.38$ & $\kappa = 1.0$ \\
Mistral-7B vs Human B & $\kappa = 0.19$ & $\kappa = 0.58$ \\
Qwen3-32B vs Human A & $\kappa = 0.69$ & $\kappa = 0.93$ \\
Qwen3-32B vs Human B & $\kappa = 0.35$ & $\kappa = 0.56$ \\
Human A vs Human B & $\kappa = 0.42$ & $\kappa = 0.69$ \\
\bottomrule
\end{tabular}
\caption{Cohen’s $\kappa$ scores for inter-annotator agreement. Binary agreement evaluates only \textit{solidarity} vs \textit{anti-solidarity}, excluding \textit{mixed} and \textit{none}.}
\label{tab:validation}
\end{table}

These results indicate moderate agreement across all classes and high reliability in avoiding \textit{fatal errors} (Binary column in Table~\ref{tab:validation}).

On the same sample of 100 statements, one annotator validated an accuracy of 89\% for the total of 342 narrative frames generated by \textit{Qwen3-32B} (i.e. every statement was assigned with 3.4 frames on average).

\section{Discussion}
\label{sec:discussion}

\subsection{High-Level Insights}
\label{sec:discussion:high-level}

Our cross-parliamentary analysis reveals key differences in how migration is discussed in the UK and the US. In the US Congress (Figure~\ref{fig:us_party_attitude}), attitudes toward migrants have become increasingly polarised along party lines over the past two decades. In contrast, UK parliamentary discourse (Figure~\ref{fig:uk_party_attitude}) remains more aligned, maintaining a more consistent gap between Labour and Conservative positions, with both shifting toward an \textit{anti-solidarity} net-attitude simultaneously.
2025 marks the most negative year in UK parliamentary discourse toward migrants in our dataset, reflecting recent political developments, such as the increasing political influence of Reform UK, the right-wing party threatening both Labour and the Conservatives.

A prominent feature of UK discourse is the prevalence of \textit{mixed} stances--statements that appear to embrace both supportive and restrictive tones simultaneously. These are far more frequent in the UK than in the US, consistent with what appears to be Britain's enduring anti-populist rhetorical norm~\cite{freeman1995antipopulist}, which discourages overt hostility toward migrants. Instead, \textit{anti-solidarity} positions are often framed as concern for fairness or responsibility, especially toward existing migrants. Yet when re-evaluated for underlying intent, most of these statements align more clearly with \textit{anti-solidarity} goals. We also observe that stronger LLMs such as \textit{Qwen3-32B} tend to identify rhetorical ambiguity more effectively in the initial annotation step than smaller models like \textit{Mistral-7B-Instruct-v0.2}, though this may be influenced by differences in model capacity, prompt sensitivity, or contextual reasoning abilities.

The analysis association between MPs' demographics and their attitude toward immigration yielded mixed results in Figure~\ref{fig:age_gender_attitude}. While female MPs persistently manifest a more \textit{solidarity} tone toward migrants, no visible association between MPs' age and net tone was seen.

\subsection{Fine-Grained Insights}
\label{sec:discussion:fine-grained}

Our frame-level analysis reveals systematic distinctions in rhetorical strategies. Looking at Figure~\ref{fig:frames-context}, \textit{solidarity} discourse often draws on \textit{international law}, \textit{cooperation}, and \textit{human rights}, while \textit{anti-solidarity} speech emphasises \textit{national law}, \textit{sovereignty}, and \textit{control}. Frames such as \textit{policy effectiveness} and \textit{public opinion} are used across both stances, suggesting shared concern for governance and legitimacy.

Figure~\ref{fig:frames_temporal} portrays partisan patterns over time. The Conservative Party has consistently focused on \textit{border control}, \textit{illegal immigration}, and \textit{national law} since the 1950s, while Labour's framing has been broader and more variable over time. Interestingly, Labour tends to emphasise \textit{border control} more strongly when in office.

Since 2020, both parties have increased references to security frames such as \textit{border control}, \textit{illegal immigration}, while cultural and economic integration frames have remained relatively stable and less emphasised. Discussions of ethnic and racial dimensions of migration peaked in the 1980s and have since declined, partially replaced by rights-based language such as \textit{migrants' human rights}, especially within Labour discourse.

Notably, the declining rate of emphasis on \textit{national law} has been mirrored by increases in the focus on \textit{international law} and \textit{human rights} over time, especially among Labour Party politicians (Figure~\ref{fig:frames_temporal_labour}). As these are percentages of a total, an increase in one is associated with decreases in others. Yet, whether we can interpret this association as causation remains an open question.

\section{Related Work}
\label{sec:related}

\subsection{Tracking Immigration Discourse}
\label{sec:related:immigration}

Research on political discourse surrounding immigration has drawn increasing attention across disciplines, particularly within computational social science. In the social media domain, \citet{mendelsohn2021frame} developed supervised models to detect immigration-related frames in tweets, showing how users’ ideology and region shape framing, and how immigration-specific frames better capture ideological variation than generic ones. Their frame taxonomy is built on prior social science literature~\cite{fryberg2012frame, Benson2013frame, boydstun2013frame}.

Closer to our domain, \citet{Card2022USDebates} conducted a large-scale analysis of 140 years of US congressional and presidential speeches. They found a long-term rise in pro-immigration sentiment from the 1940s and growing partisan polarisation more recently, with Republicans framing immigration more negatively than Democrats.

\subsection{NLP's Utility}
\label{sec:related:nlp}

Prior work has applied classic NLP techniques such as sentiment analysis, topic modelling, and supervised classification to study migration discourse. For instance, \citet{Card2022USDebates} trained a classifier to label US congressional speeches as pro-, anti-, or neutral on immigration. Similarly, \citet{mendelsohn2021frame} used supervised NLP to classify tweets into both generic and immigration-specific frames.

With the advent of LLMs, more recent work has shifted toward prompt-based approaches. \citet{kostikova2024germanparliament} used LLMs to annotate German parliamentary statements with high-level stance labels—\textit{solidarity}, \textit{anti-solidarity}, \textit{mixed}, and \textit{none}--along with hand-defined sub-level categories of \textit{group-based}, \textit{exchange-based}, \textit{empathic}, and \textit{compassionate} (\textit{anti}) \textit{solidarity}. While their work demonstrated the feasibility of LLM-based annotation for fine-grained analyses, the subtypes used for this purpose were fully predefined, subjective, and limited in scope.

\textbf{Our contribution} is both in the expansion of the scope of their application to the UK and US parliament and in the introduction of an automated approach for making a fine-grained analysis. Rather than relying on a predetermined set of granular sub-labels, we create an iterative pipeline with LLM and human feedback to discover a set of nuanced narrative frames toward migrants and subsequently AI-annotate our entire datasets of parliamentary statements accordingly.

\section{Conclusion}
\label{sec:conclusion}

This paper presents a large-scale analysis of migration discourse in the UK and US Parliaments using prompt-based annotation with open-source LLMs. Our approach combines high-level stance detection with fine-grained narrative frame extraction, offering a structured view of how migration is framed over time in the UK, across parties, and in comparison with US congressional speech.

We find that UK parliamentary discourse has become increasingly negative in tone, though it remains less polarised than US discourse. Much of this rhetoric is expressed indirectly, often through seemingly balanced statements that ultimately justify restriction. Our frame analysis further shows how control-oriented narratives have recently intensified, while deeper discussions of integration and multiculturalism have stagnated or declined.

More broadly, this work highlights the potential of lightweight LLMs and prompt engineering for political discourse analysis. Future work can expand both the scope and depth of analysis. Comparisons across discourse domains, such as parliamentary debates, news media, surveys, and social media, could reveal alignment or divergence in migration narratives. Greater granularity may be achieved by refining the frame taxonomy and incorporating speaker/author-level attributes like region or debate context to better understand rhetorical variation.

We will release the full dataset and code on GitHub in the Camera-Ready version.

\section{Limitations}
\label{sec:limitations}

Several components of our methodology involve trade-offs and approximations.

For the data collection phase, we queried a limited list of less-ambiguous keywords related to migration: ``\textit{refugee(s)}'', ``\textit{migrant(s)}'', ``\textit{immigrant(s)}'', ``\textit{asylum}'', ``\textit{migration}'', ``\textit{immigration}''. Yet, we avoided querying ambiguous terms such as ``\textit{border}'', ``\textit{settlement}'', and ``\textit{visa}'' as they can be used in contexts other than immigration to the UK (e.g. ``\textit{border}'' might frequently refer to a border between concepts rather than the UK borders). Though it is worth mentioning that this strict filtering may miss statements with implicit nuances about migration.

Our iterative pipeline for constructing the narrative frame taxonomy (Section~\ref{sec:method:association:narrative}) combines LLM-generated suggestions with expert refinement. While effective for surfacing frequent and prominent frames, this process may overlook subtle or low-frequency frames. Additionally, due to time constraints, the refinement step was only performed once; further iterations could improve the coverage and reliability of the taxonomy.

Another limitation concerns our use of chi-square tests to assess associations between frames and stance categories. The test assumes independence between variables; yet, many narrative frames tend to co-occur within the same statement, for example, ``\textit{border control}'' often appears alongside ``\textit{border sovereignty}''. This violates the independence assumption and may lead to inflated significance for correlated frames. This makes the p-values in Figure~\ref{fig:frames-context} less reliable. Yet, the effect of this is marginal, as the bar sizes already provide most of the needed insights regarding the contrast of narratives' contexts.

Finally, our analysis is limited to the UK and US parliamentary speech, which captures only the official discourse of elected representatives. As such, it may miss more informal, grassroots, or media-driven narratives that shape public attitudes toward migrants. Comparative analyses with non-parliamentary data sources could help contextualise and complement these findings.

\section{Ethical Considerations}
\label{sec:ethics}

All data used in this study are publicly available and originate from official parliamentary records produced by governments. The UK dataset was collected from the Hansard corpus, an open-access repository of parliamentary debates. The US dataset is based on the publicly released corpus compiled by \citet{Card2022USDebates}. No private or sensitive information about individuals outside of public officials was collected or analysed. As the study focuses exclusively on the discourse of elected representatives in public forums, it poses minimal risk to individual privacy.

Generative AI tools were used responsibly throughout the research. \textit{Claude Sonnet 4} was used through \textit{GitHub Copilot} for Python and \LaTeX coding and paraphrasing figure and table captions. \textit{ChatGPT} and \textit{Gemini} were utilised in the writing phase to assist with summarisation and paraphrasing of author-generated content. All Generative AI outputs were double-checked and edited by the authors.

\bibliography{references}
\appendix
\section{Appendix}
\label{sec:appendix}

\subsection{Narrative Frames Taxonomy}
\label{sec:appendix:narrative-frames}

\begin{figure*}[h]
    \centering
    \begin{subfigure}[b]{0.45\textwidth}
        \centering
        \includegraphics[width=\textwidth]{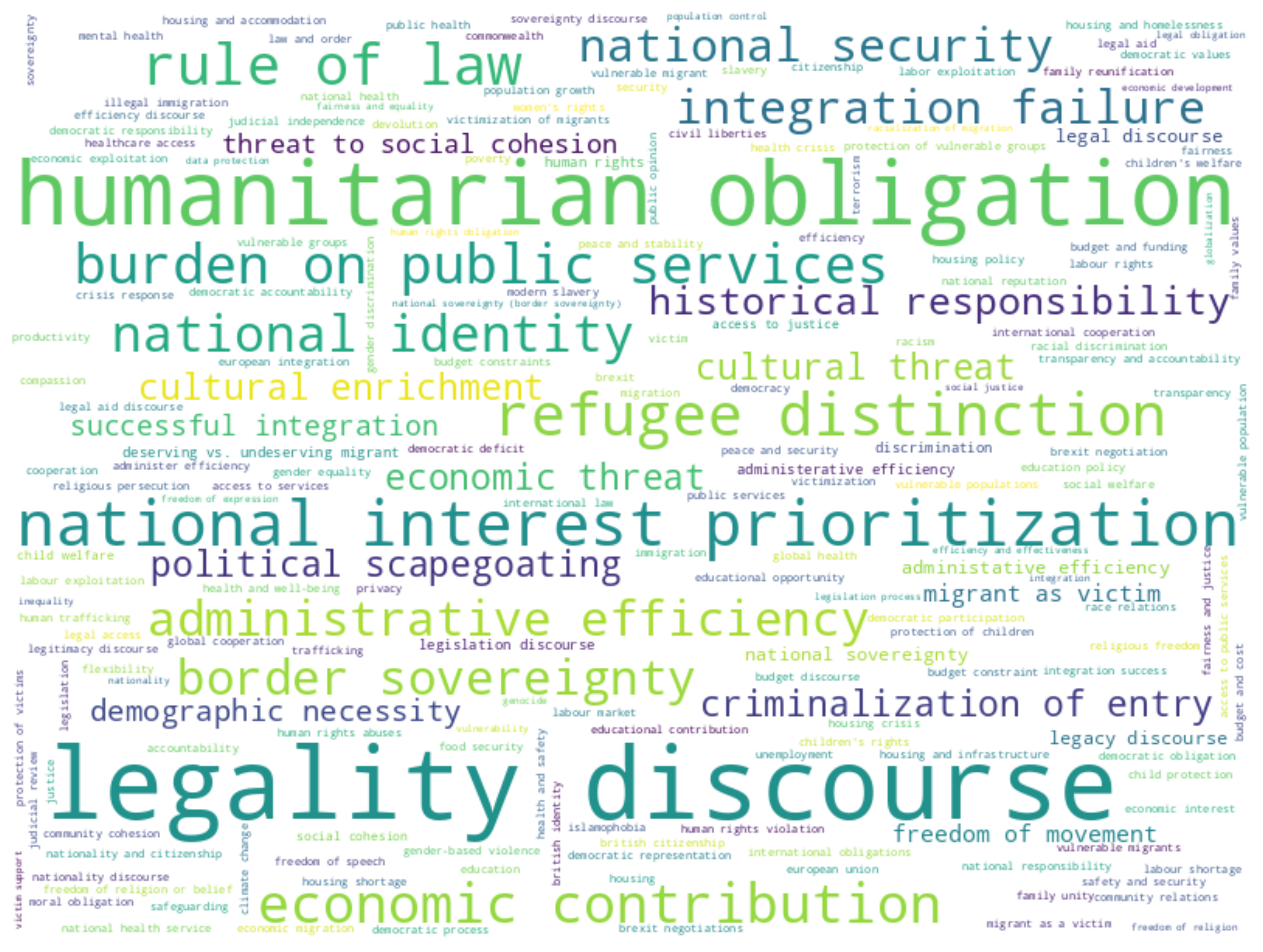}
        \caption{Stage 1: semi-fixed taxonomy (Mistral-7B)}
        \label{fig:wordcloud_stage1}
    \end{subfigure}
    \hfill
    \begin{subfigure}[b]{0.45\textwidth}
        \centering
        \includegraphics[width=\textwidth]{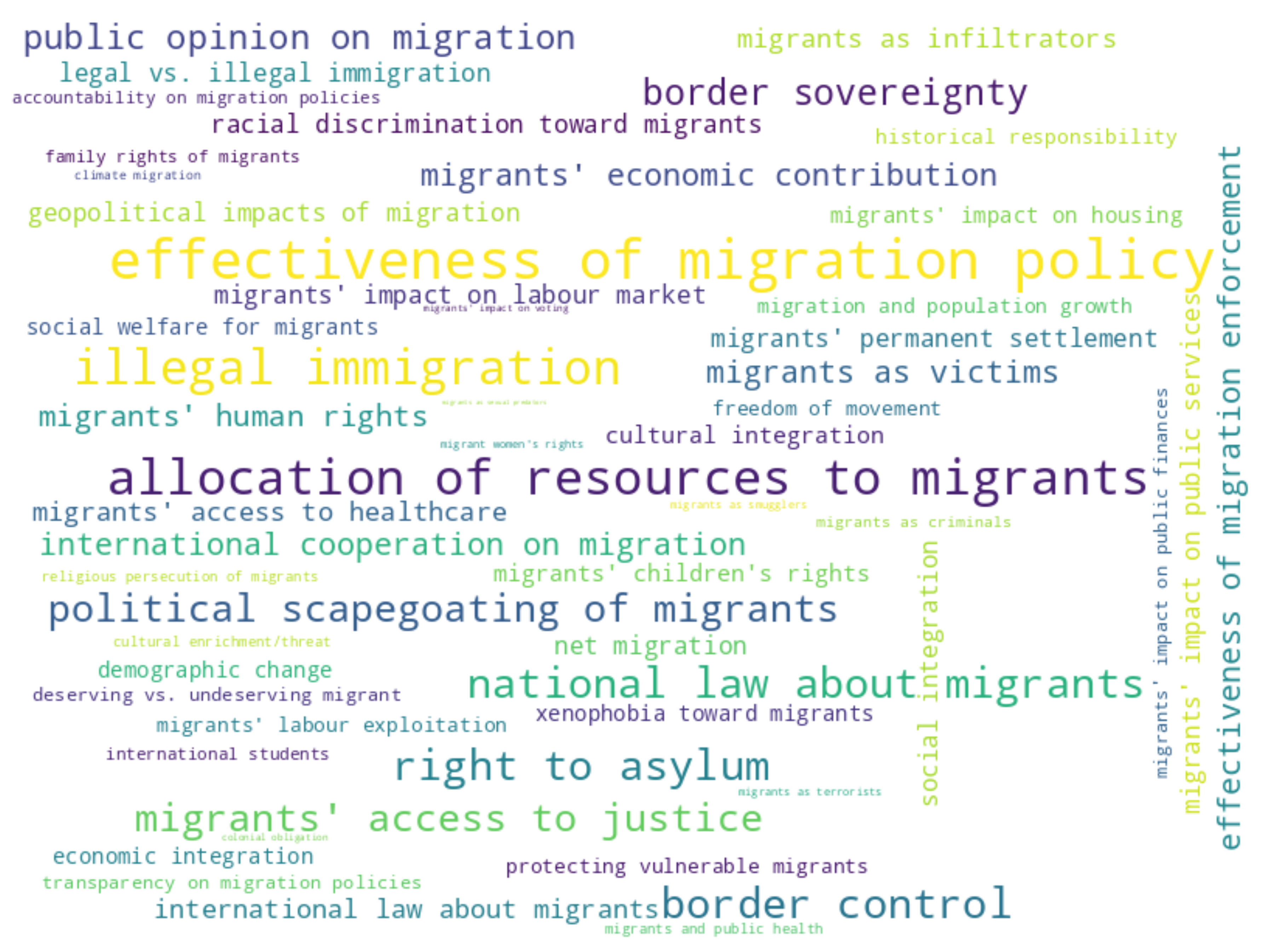}
        \caption{Stage 2: fixed taxonomy (Qwen3-32B)}
        \label{fig:wordcloud_stage3}
    \end{subfigure}
    \caption{Narrative Frame Taxonomy Progression. Word clouds show the evolution of narrative frame classification across two stages. (a) In Stage 1, a lightweight LLM (\textit{Mistral-7B-Instruct-v0.2}) was used in a semi-bounded prompt engineering setting (Table~\ref{tab:prompt_frames_unbounded}), allowing it to both assign from the prompt-provided list of frames and propose frames that are not included. The idea behind preferring a light-weight model was to allow for potential repetition of the step in future works. (b) In Stage 2, a fixed version of the taxonomy (Table~\ref{tab:prompt_frames_bounded}) was applied to the full dataset and the model was asked to only choose from the given fixed set of frames. Word cloud sizes reflect the frequency of each frame across the dataset.}
    \label{fig:wordcloud_evolution}
\end{figure*}

We portray the initial and final taxonomy of narrative frames in two wordclouds in Figure~\ref{fig:wordcloud_evolution}, where the sizes refer to the frequency of each frame being assigned to statements in the UK parliament dataset.

\subsection{Prompts}
\label{sec:appendix:prompts}

Tables~\ref{tab:prompt_relevance},~\ref{tab:prompt_solidarity},~\ref{tab:prompt_mixed},~\ref{tab:prompt_frames_unbounded}, and \ref{tab:prompt_frames_bounded} mention the full prompts used in the experiments for producing the results.

\begin{table*}[ht]
\centering
\scriptsize
\begin{tabular}{p{0.97\linewidth}}
\toprule
\textbf{Full Prompt Used for Classifying Statement Relevance to UK Migration} \\
\midrule
\texttt{You are an expert political discourse analyst. Classify whether the following statement made in the UK Parliament is RELEVANT or IRRELEVANT to migration to the UK.} \\
\texttt{\#\#\# CATEGORY DEFINITIONS} \\
\texttt{- RELEVANT: The statement is at least partially about migration to the UK (including immigration, asylum, refugees, border control, citizenship, or related policies).} \\
\texttt{- IRRELEVANT: The statement is purely procedural (e.g., scheduling, voting mechanics, parliamentary etiquette) or about topics unrelated to migration to the UK, e.g.\ it is relevant to migration to other countries.} \\
\texttt{\#\#\# INSTRUCTIONS} \\
\texttt{Return only a valid JSON output in the following format without any explanation:} \\

\texttt{\{"relevance": "<RELEVANT | IRRELEVANT>"\}}\\

\texttt{\#\#\# Text to analyze:} \\
\texttt{```\{statement\_x\}```} \\
\bottomrule
\end{tabular}
\caption{Full LLM prompt used for binary relevance classification of UK Parliament statements with respect to migration (RELEVANT vs.\ IRRELEVANT).}
\label{tab:prompt_relevance}
\end{table*}

\begin{table*}[ht]
\centering
\scriptsize
\begin{tabular}{p{0.97\linewidth}}
\toprule
\textbf{Full Prompt Used for Solidarity Annotation} \\
\midrule
\texttt{Analyze the following political statement and classify it into one of the high-level categories regarding migrants: SOLIDARITY, ANTI-SOLIDARITY, MIXED, or NONE.} \\

\texttt{If applicable, further specify by choosing the most appropriate subtype (EMPATHIC, EXCHANGE-BASED, GROUP-BASED, COMPASSIONATE) within SOLIDARITY or ANTI-SOLIDARITY.} \\

\texttt{Begin your analysis step by step:} \\

\texttt{\#\#\# HIGH-LEVEL CATEGORIES} \\
\texttt{- \textbf{SOLIDARITY}: Involves expressions that promote understanding, support, and unity with different groups or individuals (migrants), often emphasizing shared goals, compassion, mutual assistance, and empathic understanding. Consider cases with even slight expressions of solidarity.} \\
\texttt{- \textbf{ANTI-SOLIDARITY}: Entails expressions that show opposition, disregard, or exclusion towards migrants. This includes emphasizing differences, denying the need for support or assistance, highlighting unequal exchanges, and disregarding specific needs. Even slight expressions should be considered.} \\
\texttt{- \textbf{MIXED}: Characterized by both supportive and opposing expressions in the same text—e.g., recognizing migrants' rights while limiting their support. Typical indicators: conditional hospitality, selective support, policy balancing, or juxtaposition of empathy and concern with constraints or fears.} \\
\texttt{- \textbf{NONE}: Texts that do not express any solidarity or anti-solidarity toward migrants, reflecting neutrality. Absence of overt stance does not automatically mean NONE; subtle cues may still indicate a stance.} \\

\texttt{\#\#\# SUB-TYPES (only if SOLIDARITY or ANTI-SOLIDARITY)} \\

\texttt{\textbf{For SOLIDARITY:}} \\
\texttt{- \textit{EMPATHIC SOLIDARITY}: Recognizes, supports, and values differences. Applies when diversity, identity preservation, or challenges to prejudice are emphasized.} \\
\texttt{- \textit{EXCHANGE-BASED SOLIDARITY}: Based on actual or future contributions of migrants (economic, cultural, etc.). Expresses support as an investment.} \\
\texttt{- \textit{GROUP-BASED SOLIDARITY}: Driven by shared interests and goals. Emphasizes societal cohesion, rights advocacy, and anti-discrimination.} \\
\texttt{- \textit{COMPASSIONATE SOLIDARITY}: Supports marginalized/vulnerable groups with no expectation of reciprocity; acknowledges hardship and vulnerability.} \\

\texttt{\textbf{For ANTI-SOLIDARITY:}} \\
\texttt{- \textit{EMPATHIC ANTI-SOLIDARITY}: Differences should not be recognized or respected; denies group identity or diversity.} \\
\texttt{- \textit{EXCHANGE-BASED ANTI-SOLIDARITY}: Migrants are seen as taking more than they give; supports reducing help or demanding more in return.} \\
\texttt{- \textit{GROUP-BASED ANTI-SOLIDARITY}: Exclusion of migrants due to perceived incompatibility; prioritizes native interests or calls for assimilation.} \\
\texttt{- \textit{COMPASSIONATE ANTI-SOLIDARITY}: Denies help based on perception of migrants as undeserving or not truly in need; includes security or legitimacy concerns.} \\

\texttt{After your reasoning, output the label in the following JSON format:} \\

\texttt{\{"high\_level": "\textless{}SOLIDARITY \textbar{} ANTI-SOLIDARITY \textbar{} MIXED \textbar{} NONE\textgreater{}", "sub\_level": "\textless{}EMPATHIC \textbar{} EXCHANGE-BASED \textbar{} GROUP-BASED \textbar{} COMPASSIONATE \textbar{} OTHERS \textbar{} null\textgreater{}"\}}

\texttt{Only return the JSON output. Do not add explanations.} \\

\texttt{Text to analyze:} \\

\texttt{\{statement\_x\}} \\
\bottomrule
\end{tabular}
\caption{Full LLM prompt used for classifying the tone and framing of solidarity or anti-solidarity toward migrants in UK parliamentary contributions. Adopted and adapted from \citet{kostikova2024germanparliament}.}
\label{tab:prompt_solidarity}
\end{table*}

\begin{table}[ht]
\centering
\scriptsize
\begin{tabular}{p{0.97\linewidth}}
\toprule
\textbf{Full Prompt Used for Disambiguating the \textit{mixed} Stances} \\
\midrule
\texttt{Analyze the following statement from UK parliament that was **previously labeled as having a MIXED stance** toward migrants (e.g., refugees, asylum seekers, immigrants). This time, your task is to look **beyond surface-level balance or diplomatic framing** and identify the **true, final intent or policy direction** of the text toward migrants.} \\
\\
\texttt{Focus especially on:} \\
\texttt{- The **underlying policy goals or consequences** of the proposals or rhetoric,} \\
\texttt{- Whether the text **ultimately seeks to improve or restrict** the lives, rights, or social standing of migrants,} \\
\texttt{- Whether migrants are being **supported unconditionally**, or **only conditionally**, or whether support is merely rhetorical while **real-world impacts are limiting migrants benefits**.} \\
\\
\texttt{Label the **true stance** of the text with one of the following categories:} \\
\texttt{- **SOLIDARITY**: Final intent supports or uplifts migrants' rights, needs, or well-being.} \\
\texttt{- **ANTI-SOLIDARITY**: Final intent aims to reduce, restrict, undermine, or control migrant rights or support—even if framed as beneficial.} \\
\texttt{- **NONE**: No identifiable final stance. neutral or irrelevant.} \\
\\
\texttt{Output strictly the following JSON only outputting the final stance. Do not explain or comment. Do not add any introduction or text before or after the JSON.} \\
\\
\texttt{```json} \\
\texttt{\{} \\
\texttt{"final\_stance": "<SOLIDARITY | ANTI-SOLIDARITY | NONE>"} \\
\texttt{\}} \\
\texttt{```} \\
\\
\texttt{Only return the JSON output. Do not add explanations.} \\
\\
\texttt{Text to analyze:} \\
\\
\texttt{```\{statement\_x\}```} \\
\bottomrule
\end{tabular}
\caption{Full LLM prompt used for classifying the ``true intent'' of \textit{mixed} stances into \textit{solidarity}, \textit{anti-solidarity}, and \textit{none}.}
\label{tab:prompt_mixed}
\end{table}

\begin{table}[ht]
\centering
\scriptsize
\begin{tabular}{p{0.97\linewidth}}
\toprule
\textbf{Full Prompt Used for Annotating Narrative Frames} \\
\midrule
\texttt{You are an expert political discourse analyst. Given a short excerpt from a statement on migration, identify which **narrative frames** it uses to discuss migration and migrants.} \\
\\
\texttt{Select all applicable frames from the list below. If a better narrative frame is present that is **not listed**, feel free to add it using your best judgment. If the text is **off-topic, purely procedural, or contains no discernible frame**, return an empty list.} \\
\\
\texttt{Output ONLY a valid Python list of strings. No explanation, no extra text.} \\
\\
\texttt{Here is a sample list of known narrative frames:} \\
\texttt{- "Economic threat"} \\
\texttt{- "Economic contribution"} \\
\texttt{- "Cultural threat"} \\
\texttt{- "Cultural enrichment"} \\
\texttt{- "National security"} \\
\texttt{- "Humanitarian obligation"} \\
\texttt{- "Rule of law"} \\
\texttt{- "Burden on public services"} \\
\texttt{- "Demographic necessity"} \\
\texttt{- "Criminalization of entry"} \\
\texttt{- "National identity"} \\
\texttt{- "Political scapegoating"} \\
\texttt{- "Historical responsibility"} \\
\texttt{- "Integration failure"} \\
\texttt{- "Successful integration"} \\
\texttt{- "Legality discourse"} \\
\texttt{- "Refugee distinction"} \\
\texttt{- "Border sovereignty"} \\
\texttt{- "Freedom of movement"} \\
\texttt{- "Migrant as victim"} \\
\texttt{- "Deserving vs. undeserving migrant"} \\
\texttt{- "Threat to social cohesion"} \\
\texttt{- "Administrative efficiency"} \\
\texttt{- "National interest prioritization"} \\
\\
\texttt{Here is the statement, separated by triple backticks:} \\
\texttt{```\{statement\_x\}```} \\
\\
\texttt{Frames: ["Frame1", "Frame2", ...]} \\
\bottomrule
\end{tabular}
\caption{Full LLM prompt used for identifying narrative frames related to migration. The model is asked to return all applicable frames from a predefined list or add new ones if needed.}
\label{tab:prompt_frames_unbounded}
\end{table}

\begin{table}[ht]
\centering
\scriptsize
\begin{tabular}{p{0.97\linewidth}}
\toprule
\textbf{Full Prompt Used for Fine-Grained Narrative Frame Annotation} \\
\midrule
\texttt{You are an expert political discourse analyst. Given a short excerpt from a statement, identify which **narrative frames about migration** it uses to discuss migration and migrants.} \\
\\
\texttt{Select all applicable frames ONLY from the list below. If the text is **off-topic, purely procedural, or contains no discernible frame about migration**, return an empty list.} \\
\\
\texttt{Output ONLY a valid Python list of strings. No explanation, no extra text.} \\
\\
\texttt{Here is a sample list of known narrative frames:} \\
\texttt{- "protecting vulnerable migrants"}
\texttt{- "international law about migrants"}
\texttt{- "national law about migrants"}
\texttt{- "migrants' access to justice"}
\texttt{- "geopolitical impacts of migration"}
\texttt{- "migrants as infiltrators"}
\texttt{- "migrants' impact on voting"}
\texttt{- "migrants' impact on housing"} 
\texttt{- "migrants' impact on labour market"}
\texttt{- "migrants' impact on public services"}
\texttt{- "migrants' impact on public finances"}
\texttt{- "migrants and public health"}
\texttt{- "migrants' economic contribution"}
\texttt{- "migration and population growth"}
\texttt{- "net migration"}
\texttt{- "demographic change"}
\texttt{- "freedom of movement"}
\texttt{- "EU migration"}
\texttt{- "historical responsibility"}
\texttt{- "colonial obligation"}
\texttt{- "allocation of resources to migrants"}
\texttt{- "political scapegoating of migrants"}
\texttt{- "racial discrimination toward migrants"}
\texttt{- "xenophobia toward migrants"}
\texttt{- "deserving vs. undeserving migrant"}
\texttt{- "border sovereignty"}
\texttt{- "border control"}
\texttt{- "migrants' permanent settlement"}
\texttt{- "migrants and British identity"}
\texttt{- "migrants and British citizenship"}
\texttt{- "effectiveness of migration policy"}
\texttt{- "effectiveness of migration enforcement"}
\texttt{- "international cooperation on migration"}
\texttt{- "migrants' human rights"}
\texttt{- "migrants' access to healthcare"}
\texttt{- "public opinion on migration"}
\texttt{- "accountability on migration policies"}
\texttt{- "transparency on migration policies"}
\texttt{- "climate migration"}
\texttt{- "international students"}
\texttt{- "social welfare for migrants"} \\
\texttt{- "migrants' labour exploitation"}
\texttt{- "migrants as victims"}
\texttt{- "right to asylum"}
\texttt{- "religious persecution of migrants"}
\texttt{- "migrants as criminals"}
\texttt{- "migrants as sexual predators"}
\texttt{- "migrants as child abusers"}
\texttt{- "migrants as thieves"}
\texttt{- "migrants as terrorists"}
\texttt{- "migrants as smugglers"}
\texttt{- "family rights of migrants"}
\texttt{- "migrants' children's rights"} \\
\texttt{- "migrant women's rights"}
\texttt{- "LGBT migrants' rights"}
\texttt{- "cultural integration"}
\texttt{- "economic integration"}
\texttt{- "social integration"}
\texttt{- "cultural enrichment/threat"}
\texttt{- "illegal immigration"}
\texttt{- "legal vs. illegal immigration"}
\\
\texttt{Here is the statement, separated by triple backticks:} \\
\texttt{```\{statement\_x\}```} \\
\\
\texttt{Frames: ["Frame1", "Frame2", ...]} \\
\bottomrule
\end{tabular}
\caption{Full LLM prompt used for annotating migration-related narrative frames. The model selects one or more labels from a fixed list of specific frames reflecting discourse in UK parliamentary statements.}
\label{tab:prompt_frames_bounded}
\end{table}

\end{document}